\documentclass[lettersize,journal]{IEEEtran}
\usepackage{amsmath,amsfonts}
\usepackage{algorithmic}
\usepackage{algorithm}
\usepackage{array}
\usepackage[caption=false,font=normalsize,labelfont=sf,textfont=sf]{subfig}
\usepackage{textcomp}
\usepackage{stfloats}
\usepackage{url}
\usepackage{verbatim}
\usepackage{graphicx}
\usepackage{cite}
\begin{document}

\title{Transformer Based Building Boundary Reconstruction using Attraction Field Maps}

\author{Muhammad Kamran, Mohammad Moein Sheikholeslami, Andreas Wichmann, Gunho Sohn
\thanks{}
\thanks{}}



\maketitle

\begin{abstract}
In recent years, the number of remote satellites orbiting the Earth has grown significantly, streaming vast amounts of high-resolution visual data to support diverse applications across civil, public, and military domains. Among these applications, the generation and updating of spatial maps of the built environment have become critical due to the extensive coverage and detailed imagery provided by satellites. However, reconstructing spatial maps from satellite imagery is a complex computer vision task, requiring the creation of high-level object representations, such as primitives, to accurately capture the built environment. While the past decade has witnessed remarkable advancements in object detection and representation using visual data, primitives-based object representation remains a persistent challenge in computer vision. Consequently, high-quality spatial maps often rely on labor-intensive and manual processes. This paper introduces a novel deep learning methodology leveraging Graph Convolutional Networks (GCNs) to address these challenges in building footprint reconstruction. The proposed approach enhances performance by incorporating geometric regularity into building boundaries, integrating multi-scale and multi-resolution features, and embedding Attraction Field Maps into the network. These innovations provide a scalable and precise solution for automated building footprint extraction from a single satellite image, paving the way for impactful applications in urban planning, disaster management, and large-scale spatial analysis.  
\end{abstract}

\begin{IEEEkeywords}
Boundary Regularization, Attraction Field Maps, Transformers, Feature Fusion, Satellite Images, GCN.
\end{IEEEkeywords}

\section{Introduction}
\IEEEPARstart{F}{}or centuries, maps have been indispensable tools for human society, serving critical roles in navigation, planning, and understanding the world. Their applications span various fields, including urban planning, disaster response, ecological monitoring, and agricultural management. The creation of maps involves several intricate steps, such as geometric alignment to ensure accurate spatial representation, semantic labeling to add meaningful context, and vectorization to convert spatial features into abstract representations. Among these processes, this work focuses on vectorization, which transforms spatial data into simplified representations that capture objects' essential geometric and relational properties.

\par Abstract representations, such as building outlines and road networks, are fundamental to efficiently processing and analyzing spatial data. These abstractions are crucial for automating and scaling the map-making process, enabling faster storage, processing, and interpretation. However, compared to other components of cartography, the study of vector-based abstractions is less mature. While advancements in artificial intelligence (AI) and remote sensing have addressed some of the challenges in this area, achieving fully automated and reliable vectorization remains unresolved.

\par The need for automation in mapping has grown more urgent with the rapid pace of urbanization. Cities around the globe are expanding at unprecedented rates, with thousands of buildings constructed every year. This urban growth requires frequent updates to spatial maps to support urban development, infrastructure management, and environmental planning. Traditional mapping techniques, which rely heavily on manual annotation and human expertise, struggle to keep up with the demands of modern cities. Agencies like the United States Geological Survey (USGS) and the European Space Agency (ESA) have made strides in producing high-quality maps, but manual processes remain time-consuming, labor-intensive, and costly. As a result, many regions—particularly in developing countries—still lack accurate, up-to-date maps, highlighting a pressing need for scalable solutions.

\par Recent advancements in satellite technology and commercial space exploration, often referred to as the "new space era," have transformed how spatial data is collected and utilized. This era has brought unprecedented accessibility to Earth observation data, offering new insights into urban growth and land-use patterns. These innovations can potentially revolutionize map-making, enabling faster and more accurate mapping processes.

\par This paper focuses on leveraging cutting-edge deep learning techniques to automate the vectorization and abstract representation of buildings. By integrating advanced AI algorithms, geospatial analysis methods, and computationally efficient processes, this research addresses the challenges posed by rapid urbanization and the growing demand for automated mapping. Also, deep learning has emerged as a powerful tool for remote sensing applications, particularly in building footprint extraction. Our network excel at learning complex patterns from large datasets and can generalize effectively to new, unseen data. This work aims to harness these capabilities to bridge the gap between manual and automated mapping, paving the way for scalable and efficient solutions to support urbanization and global mapping efforts.

\section{Literature Review}
Recent advancements in deep learning, coupled with improvements in computational power, energy efficiency, memory capacity, and image sensor resolution, have significantly transformed the field of computer vision. These advancements have enhanced the performance and cost efficiency of vision-based applications. Deep learning models, built on multi-layer neural networks, learn hierarchical data representations, enabling them to surpass traditional techniques in tasks such as image classification, semantic and instance segmentation, and object detection. This has led to substantial improvements in accuracy and efficiency.

\par Early methods for extracting building footprints from remote sensing images can be broadly categorized into three approaches. The first approach is primitive-based, where geometric primitives such as edges and corners are extracted from images and assembled into polygons representing individual buildings. The second approach focuses on boundary-based methods, which directly learn building boundaries to generate comprehensive footprints. However, these methods face challenges when dealing with buildings of varying sizes and complex shapes. Researchers have explored the third approach, which aggregates multi-scale information to address these issues. \par Most of the research has focused on aggregating multi-scale information \cite{ref1,ref2}, while others have specifically targeted multi-scale feature extraction \cite{ref3} or designed dedicated architectures like Siamese networks \cite{ref4} and multitask networks \cite{ref5,ref6}. Although convolutional neural networks (CNNs) are effective, their spatial invariance often leads to the loss of fine details, resulting in imprecise and inconsistent building boundaries. To mitigate these limitations, advanced techniques such as signed distance transforms \cite{ref7}, frame fields \cite{ref8}, and attraction field representations \cite{ref9,ref10,ref11} have been developed to preserve intricate geometric details, particularly in complex building structures.

\par In addition to these techniques, graph models have proven effective in capturing interactions between pixels and enhancing feature representation. These models have been successfully integrated into end-to-end learning frameworks \cite{ref12, ref13}, enabling improved modeling of geometric characteristics in low-resolution imagery. Despite these advancements, regularizing building boundaries remains a key challenge in geometric learning for polygon shapes. Several segmentation models have been developed to focus on building boundaries and geometric information, often employing multi-task learning strategies. For instance, researchers have combined images and Digital Elevation Models in SegNet models to integrate additional edge and boundary predictions using a multi-task learning approach \cite{ref14}. Similarly, other studies have incorporated additional boundary losses during the training of FCN or U-Net models \cite{ref15,ref16}, generating building boundaries with enhanced regularity \cite{ref17}.

\par Modern CNN architectures have also integrated polygonal models to leverage richer geometric information. For example, the Deep Active Ray Network (DARNet) \cite{ref18} incorporates active contour models (ACM) \cite{ref19} to improve polygon contour predictions and building boundary delineation. While these methods have improved boundary regularity, segmentation models still generate building polygons at the pixel level, which often require additional smoothing and regularization. Researchers have explored various novel approaches to address this issue. For example, PSPNet \cite{ref20} is a semantic segmentation network that identifies initial building contours, while a modified PointNet predicts coordinate offsets for polygon vertices, resulting in regularized building footprints. Other efforts include fully convolutional neural networks trained with adversarial and regularized losses for boundary regularization \cite{ref21}, generative adversarial networks (GANs) that produce regularized segmentation masks \cite{ref22}, and methods that combine edge features from holistically nested edge detection with segmentation masks for regularized segmentations \cite{ref23}.

\par The earliest end-to-end deep learning-based model for building regularization was proposed by \cite{ref24}, which used a CNN to predict building polygon vertices. However, this approach suffered from fixed prediction sizes and did not fully consider the simplicity and regularity of building polygons. More advanced methods like PolyRNN \cite{ref25} and PolyRNN++ \cite{ref26} utilized recurrent neural networks (RNNs) to sequentially predict polygon vertex locations, where each prediction was influenced by prior ones. However, these methods were primarily designed for semi-automatic annotation using bounding boxes and lacked object detection capabilities. PolyMapper \cite{ref27} combined the concepts of PolyRNN and other frameworks to develop an integrated system for detecting objects and predicting sequential polygon vertices, but it struggled with complex shapes and had high computational costs due to its convolutional Long Short-Term Memory (LSTM) module \cite{ref28}. Li et al. reframed corner detection as a segmentation task and refined vertices using GCNs, while PolyWorld \cite{ref29}introduced a permutation matrix to encode vertex connectivity for final polygon generation. Multiple Networks \cite{ref30,ref31} advanced R-PolyGCN\cite{ref32} by employing oriented corners as auxiliary representations, and HiSup \cite{ref9} utilized attraction field maps for precise polygon mapping, though post-processing was needed to achieve fully regularized building boundaries.
\par Building upon these challenges and motivated to improve our baseline model, R-PolyGCN , this study presents a novel approach. Our proposed network improves upon these methods by incorporating orientation information from attraction field maps and better initializing graphs based on corner predictions. This enhancement leads to more accurate and regularized building footprints, addressing challenges in complex building extraction and regularization.

\section{Methodology}
Deep learning approaches for polygonal building segmentation can generally be categorized into two types: two-step methods and direct polygonal segmentation techniques. Two-step methods involve an initial raster segmentation, followed by post-processing steps such as vectorization or simplification. These methods often leverage auxiliary representations like frame fields or directional indicators to aid in the vectorization process. In contrast, direct polygonal segmentation techniques predict building polygons directly from the input image, bypassing intermediate rasterization steps. While direct methods streamline the process by eliminating intermediate stages, they face challenges such as missing corners, inconsistent projections, and higher computational demands. On the other hand, two-step approaches, despite being more traditional, often yield better results due to their modular and structured nature. In this work, we adopted a two-step approach, first training a decoupled object detection network on the complete image to generate high-quality bounding boxes.
\par The following sections provide a detailed discussion of the various modules in our methodology, highlighting their design, functionality, and contributions to the overall performance of the network. Each module is explained in the context of its role in addressing specific challenges and improving the accuracy and efficiency of polygonal building segmentation.
\subsection{Baseline for Object Detection}
Object detection is a critical component of our experiments, as it involves the localization of objects and the classification of their categories. In our network, the primary objective of object detection is to predict bounding boxes and their corresponding category labels for buildings. In our baseline network, PolyGCN, Mask R-CNN was used for object detection. However, it often produced bounding boxes with insufficient detail, negatively impacting the accuracy of the final polygon predictions. To address this, we explored and experimented with alternative object detection networks to enhance performance.
\par Modern object detection frameworks typically approach the task indirectly, using surrogate regression and classification problems based on proposals, anchors, or window centers. These methods often require extensive post-processing steps, such as collapsing near-duplicate predictions, designing anchor sets, and applying heuristics to assign target boxes to anchors. This reliance on handcrafted components can complicate the detection pipeline and limit efficiency.
\par To overcome these limitations, Detection Transformers (DETR)\cite{ref34} introduced a direct prediction approach that simplifies the detection process. DETR predicts all objects simultaneously and employs a set-based loss function to perform bipartite matching between predicted and ground truth objects. This end-to-end approach eliminates the need for components such as spatial anchors and non-maximum suppression, which are common in traditional object detection methods. Unlike conventional detection techniques, which rely on initial guesses (e.g., proposals in two-stage detectors or anchors in single-stage methods), DETR predicts detections directly in absolute terms with respect to the input image, thereby simplifying the pipeline and removing dependencies on handcrafted design choices. DETR can be implemented using standard components like CNNs and transformers, making it straightforward to reproduce across different frameworks.
\par Recent advancements have further refined DETR-based detectors while retaining their end-to-end simplicity. After thoroughly evaluating various DETR variants\cite{ref35, ref36,ref37}, we adopted CO-DETR, built on the Deformable-DETR framework\cite{ref38}, for its superior training efficiency and performance. CO-DETR addresses traditional one-to-one set matching limitations by introducing a collaborative hybrid assignment training scheme. This innovative approach uses one-to-many label assignments to enhance the feature learning in the encoder and attention mechanisms in the decoder. Auxiliary heads are integrated into the transformer encoder and are supervised using these versatile one-to-many label assignments, significantly improving training efficiency.
\par The CO-DETR framework adheres to the standard DETR protocol. The input image is processed through a backbone and encoder to generate latent features, which are then refined using predefined object queries in the decoder through cross-attention. The collaborative hybrid assignment training scheme enhances the encoder's feature learning and the decoder's attention learning by generating customized positive queries and optimizing label assignments. This approach significantly improves both the qualitative and quantitative performance of DETR-based detectors.

\par Given its superior performance, efficiency, and training robustness, CO-DETR was the ideal choice for our experiments. Its ability to deliver high-quality object detection results provided a solid foundation for improving downstream polygon segmentation tasks, addressing the limitations of earlier object detection approaches.
\begin{figure*}[!t]
\centering
\includegraphics[width=6.5in]{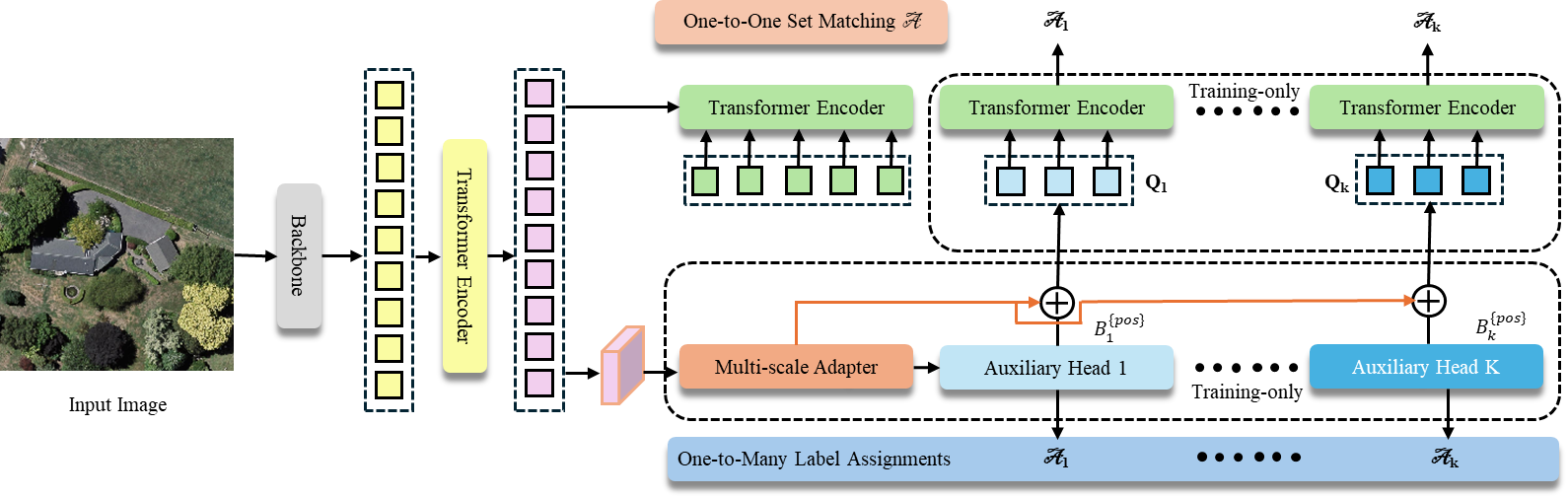}%

\label{fig:single_case}

\caption{CODETR: Baseline Network for Object Detection}
\label{fig_sim}
\end{figure*}
\subsection{Orthogonality Module}
Geometric regularization in building footprint detection and reconstruction ensures that predicted building shapes align with specific geometric constraints, reflecting real-world architectural properties such as straight edges, right angles, and symmetry. This is typically achieved by incorporating loss functions or constraints that penalize deviations from these expected properties. In remote sensing and computer vision, where satellite or aerial imagery is used to identify and outline building structures, regularization techniques are essential for enhancing the accuracy and reliability of model predictions.

\par Regularization plays a key role in reducing noise along building boundaries, ensuring that edges are smooth and continuous. This is particularly important in satellite imagery, where factors like shadows, occlusions, or imaging artifacts can introduce irregularities, leading to jagged edges in unprocessed predictions. Many urban buildings feature walls that meet at right angles, and rationalization techniques can encourage the predicted polygons to have right-angled corners where applicable, resulting in predictions that are more consistent with real-world urban architecture. By enforcing these geometric constraints, regularization improves not only the visual quality of building footprint predictions but also their quantitative performance, particularly in metrics that emphasize accurate alignment with actual shapes.

In our approach, a multi-layer Graph Convolutional Network (GCN) is used to predict vertex offsets for polygons, enabling the reconstruction of building footprints. To preserve geometric orthogonality and maintain right-angle characteristics and realistic areas, we propose a specialized orthogonality loss function. This loss function encourages the model to generate polygons with geometrically regular boundaries. By integrating orthogonality loss into the network, we achieve marginal yet meaningful gains in object detection accuracy and shape regularity, ensuring that the predicted building footprints are both accurate and visually consistent with real-world structures.
\subsection{Feature Augmentation Module}
The Feature Augmentation Module (FAM) is designed to enhance feature maps by integrating predicted vertex and edge logits into the network, effectively combining high-frequency and low-frequency features. This module plays a critical role in improving the accuracy of polygon predictions by providing enriched features to the Graph Convolutional Network (GCN), enabling more precise boundary refinement.

\par FAM addresses a key limitation observed in the RO-PolyGCN network, where low-resolution semantic features from the backbone were insufficient for capturing fine-grained details, particularly in small buildings. To overcome this, FAM augments the backbone's output with predicted edge and vertex maps, explicitly guiding the GCN on where to position vertices and edges. By incorporating high-frequency details and fusing multi-scale features, FAM preserves essential texture and color contrast, which are crucial for accurate boundary and vertex predictions, especially in complex architectural structures.

\par Integrated into our network architecture, FAM significantly enhances corner and edge detection precision, leading to more accurate building boundary reconstruction. The module utilizes a fusion block to combine high-resolution boundary features with semantic context, effectively addressing the limitations of previous low-resolution feature maps. This multi-scale integration ensures that FAM captures small building details and complex edges, providing the GCN with the enriched features it needs for improved polygon prediction. As a result, FAM is a vital component in achieving better building boundary representation and overall segmentation accuracy.
\subsection{Attraction Field Maps}
An attraction field map (AFM) represents the spatial distribution of attractive or repulsive forces within an image, generated based on features such as gradients, corners, or object boundaries. By incorporating AFMs into edge-guided feature enhancement, feature extraction can be improved by emphasizing regions with strong attraction forces. The primary purpose of an AFM is to aid in accurately detecting and delineating object boundaries within an image. This is especially beneficial in challenging scenarios, such as when noise, occlusions, or weak gradients complicate traditional edge detection methods.

\par In the context of building segmentation, AFMs can define regions of interest by modeling buildings as attractive objects, assisting in the reconstruction of building shapes. They are particularly useful for object-based image analysis, where each building is treated as an individual object. To utilize AFMs effectively, they can be concatenated with feature maps at different neural network layers, allowing the network to learn to integrate spatial attraction information with existing features.

\par Incorporating attraction field information into the loss function further enhances the network’s focus on regions with high attraction. This can be achieved by adding a penalty term to the loss function that discourages deviations from the attraction field map. However, the quality of the AFM is critical for this approach's success, necessitating the development of effective methods for generating accurate and informative attraction field maps.
\par In our network, the Attraction Field Map (AFM) provides explicit pixel-wise supervision by guiding each pixel toward the nearest boundary through a vector field. The model learns the geometric structure of building boundaries by minimizing the difference between the predicted and ground truth attraction vectors. This process encourages the network to predict vectors that accurately pull pixels toward their closest edge, enabling a more precise representation of polygonal boundaries.

Incorporating AFM loss into the backbone enhances edge detection while providing valuable orientation information. This integration significantly improves the initialization features for the Graph Convolutional Network (GCN), resulting in more accurate and refined polygon predictions
\subsection{Training}
During the training phase, ground truth bounding boxes are used to crop Regions of Interest (RoIs) from the input image. These RoIs are then passed through UResNet101, an architecture that combines the advantages of U-Net and ResNet-101 to optimize image segmentation tasks. U-Net’s symmetric encoder-decoder structure allows for precise localization by merging contextual information from the downsampling paths with detailed spatial information from the upsampling paths. ResNet-101, with its deep residual connections, improves training stability by addressing vanishing gradient issues. By incorporating ResNet-101 as the encoder in U-Net, UResNet101 is able to capture complex features while preserving localization accuracy. This combination produces high-resolution segmentation maps, which are particularly effective for building footprint extraction. The output features from UResNet101 match the dimensions of the RoIs, ensuring localized and precise features for downstream predictions.
\begin{figure*}[!t]
\centering
\includegraphics[width=6.5in]{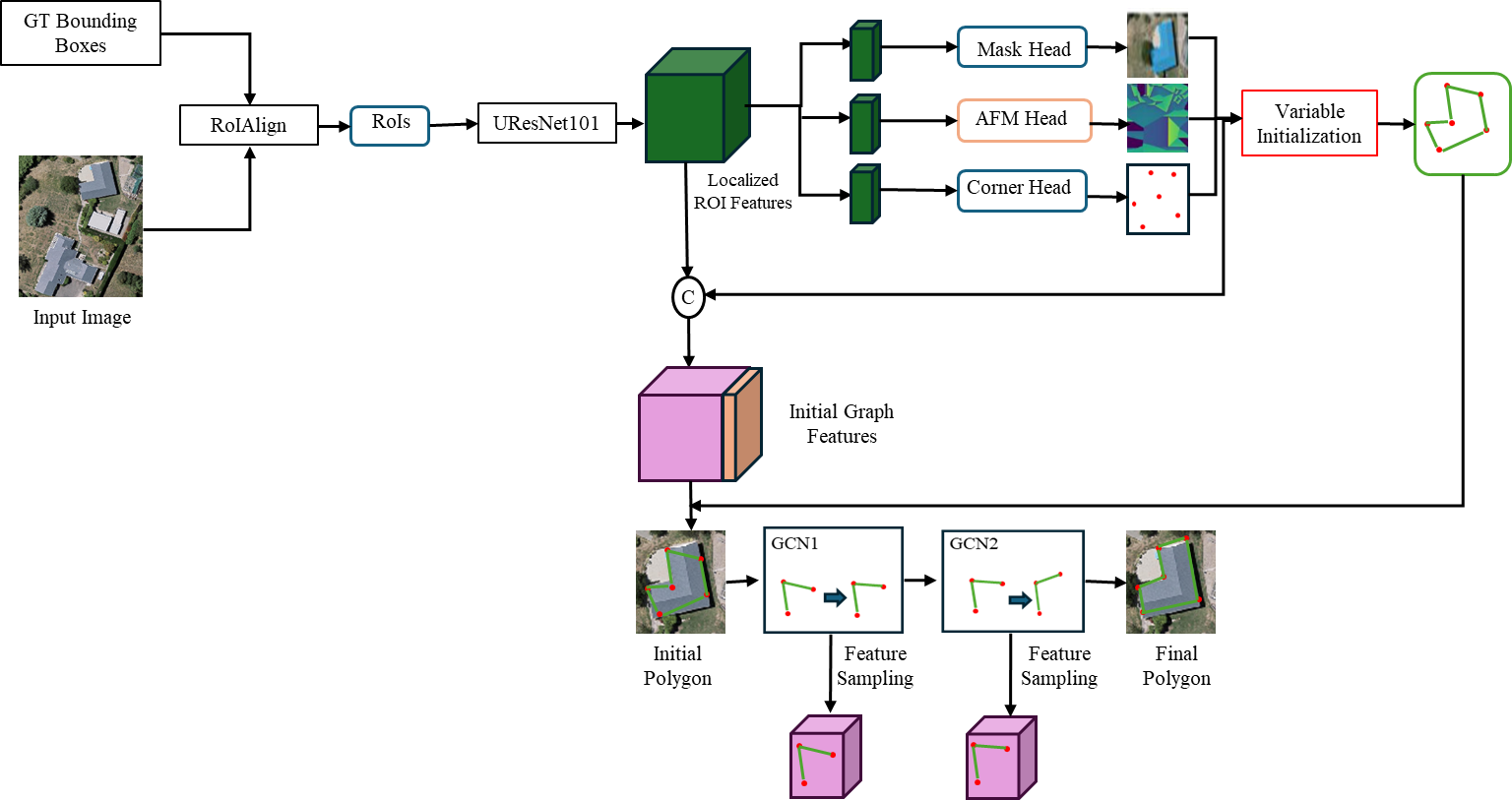}%

\label{fig:single_case}

\caption{Training}
\label{fig_sim}
\end{figure*}

\par From these feature maps, the network employs three prediction heads: the Segmentation Head, Attraction Field Map (AFM) Head, and Corners Head, similar to the setup in PolyAttractNet. These heads generate highly localized predictions, which are subsequently processed by the Graph Convolutional Network (GCN) for producing accurate building polygon predictions. To further enhance the features, boundary masks generated from the RoI features are concatenated with the original RoI feature map.

\par For polygon initialization, dynamic points are used, leveraging predictions from both masks and corners. Corners are classified into two types: convex and concave. Mask contours are extracted and simplified, followed by applying thresholds to identify convex and concave corners. A non-maximum suppression (NMS) technique is applied to refine the extracted corners. These corners are then combined with the mask contours, and the closest points on the contour to each corner are identified to ensure proper border connectivity. This process generates an initial polygon. Additionally, any missing corners that were not included in the initial polygons are identified from the mask contour and added to their appropriate positions, resulting in a refined initial polygon ready for further optimization by the GCN.

\par The Graph Convolutional Network (GCN) is utilized to iteratively refine the polygon vertices. Starting from the initial polygon, a three-step GCN process is employed to progressively improve vertex positions. In the first step, the graph features are processed by the GCN to compute vertex offsets, which adjust the positions of the vertices. The updated vertex positions are then used to interpolate new graph features, which are processed by a second GCN to compute further offsets, refining the vertices further. This iterative process is repeated for three steps, continuously improving the accuracy of the polygon predictions. To ensure that the final polygons are geometrically regularized, orthogonality loss is applied, resulting in sharp, well-aligned edges and precise building boundaries. This comprehensive approach ensures robust and accurate building footprint extraction.

\subsection{Inference}
During the inference phase, the overall network design is similar to that of the training phase, with the primary difference being the input to the UResNet101 backbone. In inference, the input features are derived from bounding boxes predicted by a transformer-based object detection network, CO-DETR. After extensive experimentation, CO-DETR was selected as the optimal detection network due to its ability to produce high-quality bounding boxes. These bounding boxes are used to generate Regions of Interest (RoIs) from the input image.
\par The output features are then concatenated with predictions from the Mask Head, AFM Head, and Corners Head, following the same process as in the training phase. Polygons are initialized dynamically using features from the mask and corner predictions, and a multi-step Graph Convolutional Network (GCN) is applied to iteratively refine the boundaries. The inclusion of orthogonality loss during refinement ensures that the final polygons are geometrically regularized and accurate. This two-step approach emphasizes the critical importance of high-quality object detection for effective polygonal segmentation. By enhancing both raster-based and vector-based metrics, Decoupled-PolyGCN significantly improves the accuracy, regularity, and reliability of building footprint predictions, making it a robust solution for real-world applications.

\begin{figure*}[!t]
\centering
\includegraphics[width=6.5in]{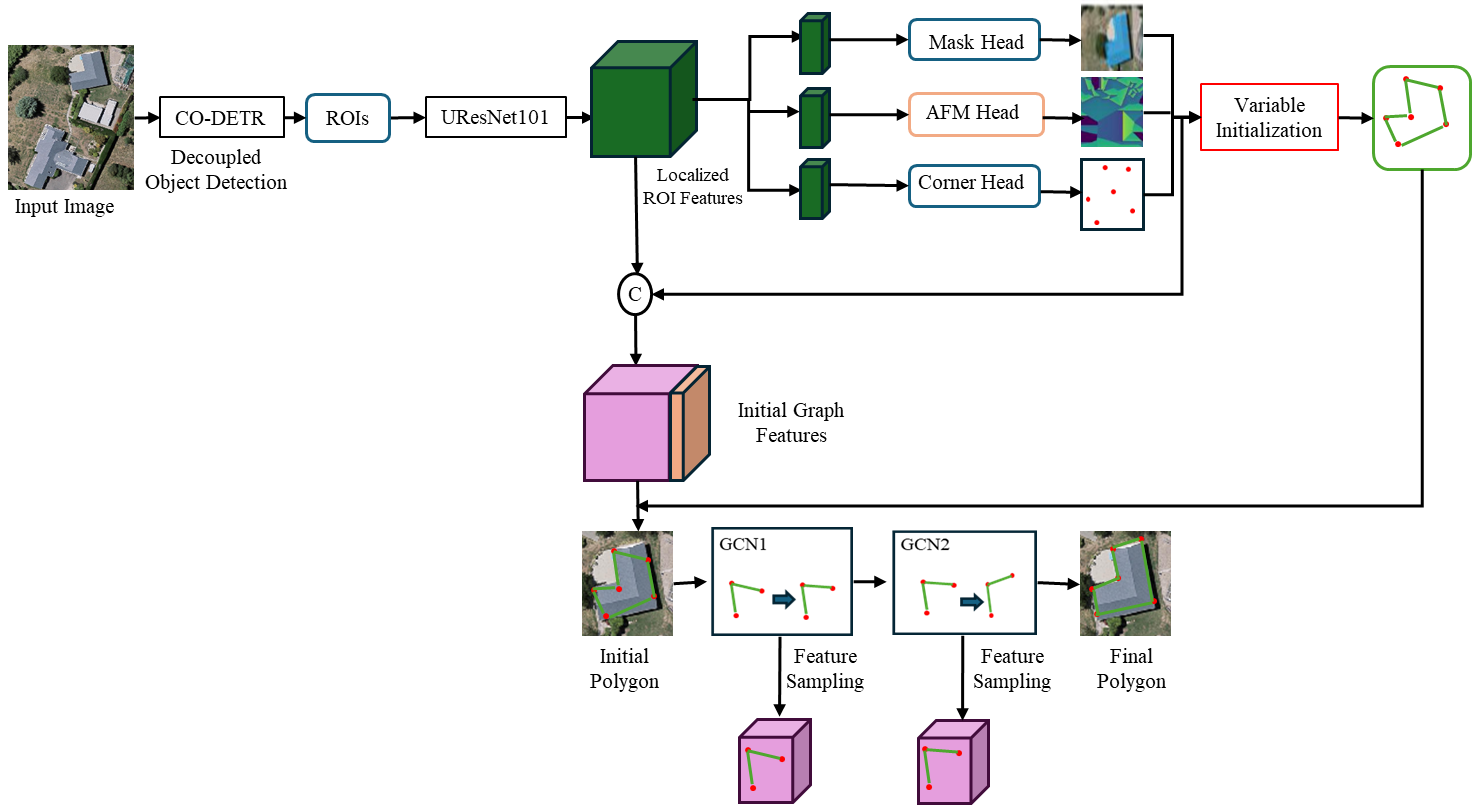}%

\label{fig:single_case}

\caption{Inference}
\label{fig_sim}
\end{figure*}

\section{Loss Function}
Our network employs a multi-loss training strategy, as discussed in previous chapters, to optimize different components of the model. Cross-entropy loss is widely used for instance segmentation tasks, particularly for mask prediction, to quantify the difference between the predicted mask probabilities and the ground truth. For binary mask prediction, which classifies each pixel as either foreground or background, we use the binary cross-entropy (BCE) loss.
\begin{equation} L_{mask} (p(y))= - (y log(p(y))+ (1-y)log(1  p(y))) \end{equation}
For the building corners, which are categorized into convex and concave corners, we also apply cross-entropy loss to optimize their predictions.
\newline The Attraction Field Map (AFM) loss is designed to minimize the discrepancy between the predicted attraction vectors and the ground truth attraction vectors. If N is the total number of pixels in the image, P represents the set of all pixels in the image.  $\tilde{v}$ is the predicted attraction vector at pixel p.  $v_p$ is the ground truth attraction vector at pixel p. Using the $L1$ loss, this is formulated as. 
\begin{equation} L_{\text{AFM}} = \frac{1}{N} \sum_{p \in P} \left\| \tilde{v} - v_p \right\|^2 \end{equation}
\par The features from the mask and corners are used to construct an initial polygon with a dynamic size, which is subsequently passed to the Graph Convolutional Network (GCN) for further refinement. The loss function for the GCN is designed to optimize polygon vertex prediction by minimizing the difference between the predicted vertex positions $p^{pre}$ and the ground truth vertex positions $p^{gt}$. This is calculated using the geometric $L1$ distance.
\begin{equation} L_1(p^{\text{pre}}, p^{\text{gt}}) = \sum_{i=0}^N \left( |x_i^{\text{pre}} - x_i^{\text{gt}}| + |y_i^{\text{pre}} - y_i^{\text{gt}}| \right) \end{equation}
To enforce geometric regularity and ensure that building boundaries retain their orthogonal properties, we introduce an orthogonality loss. This loss helps maintain right-angle characteristics and improves the shape regularity of building polygons. The orthogonality loss is defined as
\begin{equation} L_{ortho} = \frac{1}{N} \sum_{j=1}^N L(P_j) \end{equation}
For training, we combined different loss function weights and conducted extensive hyperparameter tuning to achieve optimal network convergence. Hyperparameter tuning is crucial in deep learning as it directly impacts learning, generalization, and stability. Balancing the weights of loss components, such as mask loss, AFM loss, vertex prediction loss, and orthogonality loss, was key to ensuring no single task dominated the optimization process, which could degrade performance. For example, overemphasizing mask loss could lead to well-segmented masks but poorly regularized polygons, while underweighting it could produce coarse masks, hindering downstream tasks. Tuning involved systematically experimenting with parameters like learning rates, batch sizes, and optimizer settings. This helped the network capture fine-grained boundary details while maintaining geometric regularity. Learning rates were particularly critical, as rates that were too high or low could hinder convergence. Similarly, batch size affected stability and computational efficiency, with smaller batches improving generalization but increasing noise.
\par Through this process, we stabilized the training and optimized various aspects of building footprint extraction, including mask prediction, corner detection, polygon refinement, and geometric regularity. This comprehensive approach significantly improved the network's performance, resulting in more accurate segmentation and precise polygonal predictions for building footprint reconstruction.
\section{Experiments and Results}

\subsection{Dataset}
The primary dataset utilized in this study is SpaceNet-2, an open-access dataset originally created for the building extraction challenge at the DeepGlobe workshop held during CVPR 2018. This dataset comprises high-resolution satellite imagery with precise building boundary annotations. Covering both urban and suburban areas, it features data from four cities across four continents—Las Vegas, Paris, Shanghai, and Khartoum—ensuring a diverse range of roof appearances across various geographic regions. The images were captured by the DigitalGlobe WorldView-3 satellite in GeoTIFF format, with a resolution of 30 centimeters per pixel. Each image measures 650x650 pixels, corresponding to a ground area of 200x200 meters. The dataset includes 24,586 labeled satellite images containing 302,701 building footprint polygons, all meticulously annotated and provided in GeoJSON format. The experimental data were obtained through Amazon Web Services (AWS) under a license from the DeepGlobe workshop. For our research, we specifically selected Pan-sharpened RGB images from the available satellite imagery types.

\par In addition to SpaceNet-2, we incorporated the WHU Building Dataset as a secondary dataset for training and testing our network. This dataset comprises approximately 220,000 annotated building footprints derived from aerial imagery. The images have dimensions of 300x300 pixels, with an impressive spatial resolution of 0.075 meters per pixel, covering a total area of 450 square kilometers in Christchurch, New Zealand. The dataset was constructed using imagery captured from multiple cities worldwide through remote sensing platforms such as QuickBird, WorldView series, IKONOS, and ZY-3. Together, these datasets provide a robust and diverse foundation for evaluating and improving building footprint extraction methods.

\subsection{Performance Metrics}
\subsubsection{Raster-Based Metrics}
Raster-based building evaluation metrics are quantitative tools used to evaluate the performance of building extraction models by comparing predicted building masks with ground truth masks at the pixel level. These metrics assess the model's ability to accurately classify pixels within raster images, offering a detailed understanding of pixel-wise classification accuracy. They are particularly useful for analyzing how well the model captures the extent and shape of buildings.
\newline \textbf{Precision} evaluates the accuracy of the model’s positive predictions, indicating the proportion of correctly identified building pixels out of all pixels classified as buildings. A high precision score implies that the model has fewer false positives, making it especially valuable in applications where incorrect detections are costly. Precision is calculated as:
\begin{equation} Precision=\frac{TP}{TP+FP} \end{equation}
where TP (True Positives) represents correctly identified building pixels and FP (False Positives) denotes pixels incorrectly classified as buildings.
\newline\textbf{Recall} measures the model's ability to identify all relevant building pixels, showing the proportion of actual building pixels that were correctly detected. High recall indicates that the model has fewer false negatives, which is critical in scenarios where missing detections can have significant consequences. Recall is calculated as:
\begin{equation} Recall=\frac{TP}{TP+FN}\end{equation}
where FN (False Negatives) represents building pixels that were missed by the model.
Together, precision and recall provide a comprehensive evaluation of a model’s performance. While precision reflects the accuracy of positive predictions, recall indicates the model’s effectiveness in detecting all relevant objects. Balancing these metrics is essential for achieving robust building extraction performance.
\subsubsection{Vector-Based Metrics}
Vector-based building evaluation metrics evaluate the performance of building extraction models by measuring the geometric accuracy of predicted building polygons rather than pixel-level masks. These metrics focus on assessing the alignment, shape, and structural accuracy of predicted polygons in comparison to ground truth polygons, making them highly relevant for applications requiring precise building boundaries and detailed shapes, such as urban planning and mapping. By emphasizing boundary precision, shape preservation, and structural fidelity, vector-based metrics provide a more refined analysis of model performance. They offer valuable insights into how well the predicted polygons capture the true geometry, edges, and structures of buildings, particularly in complex urban environments.
\newline \textbf{PoLiS} (Polygonal Line String Similarity)  is a vector-based metric specifically designed to evaluate the similarity between predicted and ground truth building polygons, with a focus on shape and structural alignment rather than pixel-level accuracy. It is particularly suited for assessing the geometric precision of building boundaries in building extraction tasks.

\par The PoLiS metric calculates the shortest distance from each point on the predicted polygon to the closest point on the ground truth polygon, and vice versa. This two-way distance calculation ensures a balanced evaluation, accounting for differences in complexity between the two polygons. By measuring alignment and shape similarity, PoLiS captures discrepancies in spatial arrangement and structural detail, making it highly effective for applications requiring precise reproduction of building outlines.

\par PoLiS evaluates both the positional alignment and boundary similarity of the polygons, taking into account the full perimeters of both the predicted and ground truth shapes. This comprehensive approach ensures a nuanced assessment of geometric accuracy, emphasizing the importance of accurate boundary representation in building extraction tasks. It is calculated as the average distance between the matched vertices in polygons P and Q:

\begin{multline} 
\text{PoLiS}(P, Q) = \frac{1}{2N_P} \sum_{p_j \in P} \min_{q \in Q} \left\| p_j - q \right\| \\
+ \frac{1}{2N_Q} \sum_{q_k \in Q} \min_{p \in P} \left\| q_k - p \right\|
\end{multline}
\textbf{MaxTangent} (Maximum Tangent Angle Error) is a vector-based evaluation metric designed to assess angular alignment between the boundaries of predicted and ground truth building polygons in building extraction tasks. By analyzing the tangent angles of corresponding boundary segments, MaxTangent evaluates how well the predicted polygon aligns with the true geometry, particularly in terms of edge orientation and corner accuracy.

\par The metric calculates the absolute difference in tangent angles for each boundary segment, capturing deviations in orientation. The maximum angular difference across all segments is recorded as the MaxTangent value, representing the worst-case misalignment. This emphasis on boundary orientation and edge alignment makes MaxTangent particularly effective for evaluating structural fidelity, as it is highly sensitive to corner preservation, edge alignment, and overall boundary regularity.

\par Unlike metrics based on area or distance, MaxTangent focuses specifically on angular discrepancies, identifying misaligned edges and irregularities at sharp corners. This precision in measuring angle accuracy is especially valuable for ensuring that models preserve architectural features and the structural integrity of buildings with well-defined edges and geometric shapes.
\par For each edge in P and Q, calculate the tangent vector as the difference between two consecutive vertices: The tangent error at each point pi is given by the distance between the tangent vectors of $p_i$ and its closest point q. The MaxTangent metric is then the maximum of these distances over all vertices $p_i$  in P.

\subsection{Results}
\begin{figure}[!t]
\centering
\includegraphics[width=3.45 in]{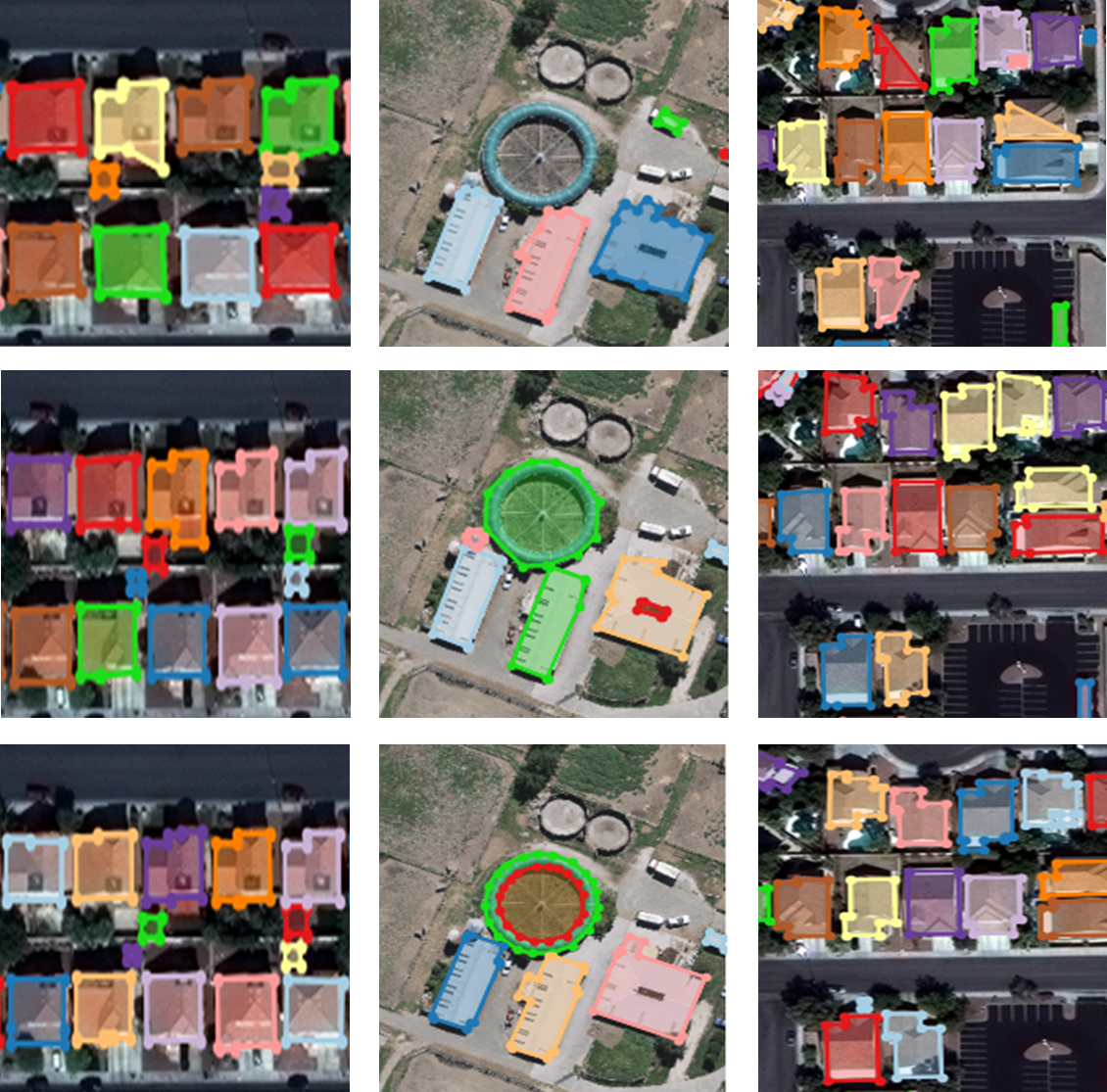}
\caption{Comparison of Results with SOTA Model, HiSup. The top row indicates prediction from the HiSup Model, the middle row displays the boundary polygons predicted by DeCoupled-
PolyGCN, and the last row is the ground truth.}
\label{fig_1}
\end{figure}

Figure \ref{fig_1} presents a comparison between our proposed Decoupled-PolyGCN and the SOTA model HiSup [109]. The results clearly show that DeCoupled-PolyGCN significantly outperforms HiSup, delivering superior performance across various scenarios. The HiSup model struggles with complex-shaped buildings and fails to effectively distinguish closely located structures. DeCoupled-PolyGCN, on the other hand, proves to be more robust and versatile, performing consistently well under diverse conditions and addressing these challenges with greater accuracy and precision. We thoroughly analyzed and refined our network to ensure it performs robustly across diverse scenarios, addressing various challenges presented by different datasets and building characteristics.
\par To evaluate the performance of our proposed models, we conducted a comprehensive quantitative
analysis using both raster-based and vector-based metrics. This dual evaluation approach
highlights the effectiveness of our models not only in pixel-level segmentation but also in
accurately capturing the geometric regularity of building footprints. Our results demonstrate
the consistent progression of performance across the proposed models, with each new module
contributing to improved accuracy and boundary regularization.
\par Table \ref{tab:methods_overview_WHU} presents the performance of our models on the WHU dataset, highlighting
consistent improvements across all evaluated metrics. Each successive module contributes
to enhancing the network’s overall effectiveness. Our final model, Decoupled-PolyGCN,
achieves the best results, delivering highly regularized building boundaries and setting new
benchmarks for state-of-the-art performance. There is a 20\% improvement in AP and 17\% in
AR in DeCoupled-PolyGCN as compared to the baseline model, R-PolyGCN. Notably, lower
values of PoLiS and MTA indicate better performance. The results clearly show a progressive
decline in these values with each model iteration, highlighting the steady advancements in
our network’s geometric capabilities capabilities.
\begin{table}[h!]
    \centering
    \caption[Overview of Results on WHU Dataset]{Overview of Results on WHU Dataset. Lower values of POLIS and MTA are better.}
    \begin{tabular}{|l|c|c|c|c|}\hline
			\textbf{Networks}&\textbf{AP(\%)}&\textbf{AR(\%)}&\textbf{PoLiS $\downarrow$\ } & \textbf{MTA $\downarrow$\ }\\ \hline
			 R-PolyGCN&45.7&57.5&2.58 & 45.23\\ 
             \hline
			 RO-PolyGCN&46.1&58.3&1.96&43.88\\ 
             \hline
			 FAE-PolyGCN&53.0&61.4&1.90&43.35\\ 
             \hline
			 AT-PolyGCN&55.3&62.5&1.46&42.16\\ 
             \hline
             DeCoupled-PolyGCN& \textbf{65.7}&\textbf{74.5}&\textbf{1.22}&\textbf{37.95} \\ \hline
    \end{tabular}
    \label{tab:methods_overview_WHU}
\end{table}
Similarly, Table \ref{tab:methods_overview_Vegas} shows the quantitative results for our models on the SpaceNet-2
(Vegas) dataset. As with the WHU dataset, our models consistently improve with each
added module. DeCoupled-PolyGCN, in particular, stands out by delivering highly accurate
segmentation results and regularized polygonal boundaries, making it our most robust and
efficient model. There is a 15\% improvement in AP and 12\% in AR in DeCoupled-PolyGCN
as compared to the baseline model, R-PolyGCN. As discussed earlier, lower values of PoLiS
and MTA indicate better performance. The results clearly show a continuous decrease in these
values with each model iteration. This showcases the improvement in geometric capabilities
of our network.
\begin{table}[h!]
    \centering
    \caption[Overview of Results on SpaceNet-2 Dataset]{Overview of Results on SpaceNet-2 Dataset. Lower values of POLIS and MTA are better.}
    \begin{tabular}{|l|c|c|c|c|}\hline
			\textbf{Networks}&\textbf{AP(\%)}&\textbf{AR(\%)}&\textbf{PoLiS $\downarrow$\ } & \textbf{MTA $\downarrow$\ }\\ \hline
			 R-PolyGCN&42.1&53.5&3.51 & 44.65\\ 
             \hline
			 RO-PolyGCN&44.5&55.1&3.25&41.27\\ 
             \hline
			 FAE-PolyGCN&50.4&57.0&3.20&41.19\\ 
             \hline
			 AT-PolyGCN&52.0&58.2&3.12&40.35\\ 
             \hline
             DeCoupled-PolyGCN& \textbf{57.8}&\textbf{65.5}&\textbf{2.78}&\textbf{36.51} \\ \hline
    \end{tabular}
    \label{tab:methods_overview_Vegas}
\end{table}
\par We conducted a comprehensive evaluation of our models alongside a comparison with
several SOTA methods from the literature. For this, we selected both two-step and direct
polygonal segmentation approaches to compare against our model. Table \ref{tab:methods_overview_competing_spaceNet-2} provides a
detailed overview, showcasing how our Decoupled-PolyGCN model surpasses existing methods.
Notably, our model achieves a 5\% improvement in AP and a 7\% increase in AR, representing
a significant performance improvement. This comparison clearly shows the advancements
made by our model in terms of both accuracy and geometric regularity. Furthermore, it
highlights that our approach consistently outperforms both two-step and direct polygonal
segmentation SOTA methods, demonstrating its effectiveness and robustness.
\begin{table}[h!]
    \centering
    \caption{Comparison of DeCoupled-PolyGCN with competing methods on SpaceNet-2 Dataset}
    \begin{tabular}{|c|c|c|}
        \hline
        \textbf{Networks} & \textbf{AP(\%)} & \textbf{AR(\%)} \\ \hline
        MaskRCNN & 43.2 &52.5\\
        \hline
        R-PolyGCN & 42.1 &53.5\\
        \hline
        PolyMapper & 49.6 &55.0\\
        \hline
        FrameField & 52.1 &57.3\\
        \hline
        HiSup  & 52.5 & 58.3\\
        \hline
        Ours(DeCoupled-PolyGCN) & \textbf{57.8}&\textbf{65.5}\\
        \hline
    \end{tabular}
    \label{tab:methods_overview_competing_spaceNet-2}
\end{table}
\par Alongside the Vegas dataset, we carried out experiments on the WHU dataset. We
compared our results with the two-step and direct polygonal methods. The comparison is
shown in table \ref{tab:methods_overview_competing_WHU} , which clearly demonstrates our network to be superior to the SOTA
methods. It improves 6\% in AP and 10\% in AR.
\begin{table}[h!] 
    \centering
    \caption{Comparison of DeCoupled-PolyGCN with competing methods on WHU Dataset}
    \begin{tabular}{|c|c|c|}
        \hline
        \textbf{Networks} & \textbf{AP(\%)} & \textbf{AR(\%)} \\ \hline
        MaskRCNN  & 45.9 &55.4\\
        \hline
        R-PolyGCN  & 45.7 &57.5\\
        \hline
        HiSup & 59.9 & 64.0\\
        \hline
        Ours(DeCoupled-PolyGCN) & \textbf{65.7}&\textbf{74.5}\\
        \hline
    \end{tabular}
    \label{tab:methods_overview_competing_WHU}
\end{table}
\subsection{Discussions}
We began our research by utilizing R-PolyGCN as the baseline model, systematically building upon it to improve network performance. Over time, multiple modules were added to address challenges and enhance the network's capability. Continuous analysis of the network was conducted under diverse experimental settings. The datasets chosen for our experiments were highly diverse, featuring buildings of varying sizes and shapes, with scenes ranging from dense rural areas to crowded urban environments. These presented significant challenges, requiring a robust solution capable of generalizing across all conditions.
\newline \textbf{Size-Based Analysis:} The performance of our network varied significantly based on the size of the buildings. Small buildings posed notable challenges for segmentation due to their intricate details, often leading to lower average precision (AP) scores. Most errors occurred with buildings ranging from 50 to 200 square pixels, where structures were either entirely missed or incorrectly merged with adjacent ones. Medium-sized buildings, comprising nearly 70\% of the dataset, demonstrated the best overall performance across all models. In contrast, larger buildings, though less frequent in the dataset, revealed a persistent performance gap.

\par
\begin{figure}[!t]
\centering
\includegraphics[width=3.45 in]{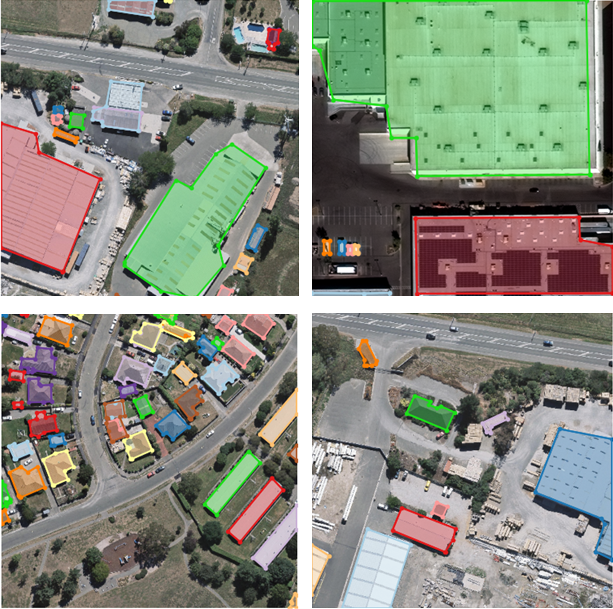}
\caption{Size Based Analysis}
\label{fig_2}
\end{figure}

 Our baseline model struggled significantly with detecting small buildings. To address this, we introduced the Feature Augmentation Module (FAM), which improved high-resolution feature representation. This enhancement led to an 8\% increase in AP and a 5\% increase in average recall (AR) for smaller buildings. However, larger buildings remained challenging for the network. To mitigate this, the Attraction Field Maps (AFM) module was integrated, encoding orientation information to enhance the segmentation of large structures. While AFMs had minimal impact on small and medium-sized buildings, they substantially improved performance for large buildings, yielding a 14\% improvement in AP and a 10\% increase in AR. To further refine performance, a decoupled object detection strategy was implemented. This approach brought significant advancements for both small and large buildings. The dynamic graph initialization in this model effectively reduced missed corners, resulting in more precise segmentation. For small buildings, our final model achieved a 3\% improvement in AP and an 8\% increase in AR. Similarly, for large buildings, it delivered a 13\% boost in AP and a 12\% increase in AR. The results of the final model, DeCoupled-PolyGCN, are visually depicted in Figure \ref{fig_2}, demonstrating the network's capability to segment buildings of varying sizes with enhanced precision and accuracy. These advancements highlight the effectiveness of our approach in addressing the unique challenges associated with diverse building sizes.
\newline \textbf{Vertex-Based Analysis:} The datasets used in this study predominantly consisted of buildings with 5–8 or 9–16 vertices, with fewer examples falling into the categories of $leq4$ or >32 vertices. For our experimental evaluation, we categorized the buildings into four vertex-based groups: ($\leq 4$, $5-8$, $9-16$, and $>16$) vertices. Initially, we adopted a fixed vertex count of 16 for graph initialization. While this approach performed well for buildings with 9–16 vertices, it often led to distortions for other categories due to over-sampling or under-sampling, which adversely affected shape accuracy. For instance, buildings with $leq4$ vertices were particularly vulnerable to IoU degradation when key corners were missed. Similarly, smaller buildings with $>16$ vertices suffered from inaccuracies in polygon predictions, as the fixed vertex count failed to capture their geometric intricacies. On the other hand, buildings with very high vertex counts required careful handling to prevent cumulative vertex placement errors, which could increase PoLiS values. 
\par To address these issues, we experimented with various fixed vertex counts, including 8, 32, and 96, as these values are commonly used in different scenarios. While a higher count of 96 vertices performed well for complex structures, a fixed count of 16 vertices emerged as the optimal choice for general cases, offering a balance between accuracy and computational efficiency. Our final model, Decoupled-PolyGCN, introduced a dynamic graph initialization strategy. This approach adapts the number of vertices based on the geometric complexity of each building, resulting in polygons with more accurate and regularized boundaries. This adaptability significantly enhanced both quantitative and qualitative performance.  The improvements are particularly evident in the WHU dataset. For buildings with less than or equal to 4 vertices, Decoupled-PolyGCN achieved a 10\% increase in Average Precision (AP) and Average Recall (AR) compared to the baseline. For buildings with $>16$ vertices, AR showed a substantial improvement of 24\%. On the SpaceNet-2 dataset, the enhancements were even more pronounced, with AP increasing by 30\% and AR improving by 12\% for buildings with $>16$ vertices. 

By adapting to the complexity and shape of each building, Decoupled-PolyGCN ensures higher accuracy and efficiency in polygon predictions. This adaptability enables the model to consistently deliver improved performance across varying building complexities and geometries, setting a new standard for building footprint extraction.
\newline \textbf{Proximity-Based Analysis:} Dense urban environments, characterized by closely spaced buildings, posed a significant challenge in segmentation tasks due to overlapping boundaries and shared edges between adjacent structures. This often led to segmentation errors where multiple buildings were incorrectly merged into a single structure, particularly in earlier models that lacked the precision for fine-grained boundary delineation. Such limitations negatively impacted the overall performance of the network, as these models struggled to accurately capture the distinct boundaries of buildings in dense settings.

\par Our final model, Decoupled-PolyGCN, addressed these challenges by leveraging the CO-DETR detection network, which significantly improved the accuracy of bounding box predictions. These enhanced bounding boxes provided a stronger foundation for the Graph Convolutional Network (GCN) by supplying high-quality initialization features. As a result, the network could better distinguish closely spaced buildings, effectively reducing instances of merged boundaries. This improvement in graph initialization directly contributed to more precise and well-separated polygon predictions, even in densely packed urban environments.

\par
\begin{figure}[!t]
\centering
\includegraphics[width=3.45 in]{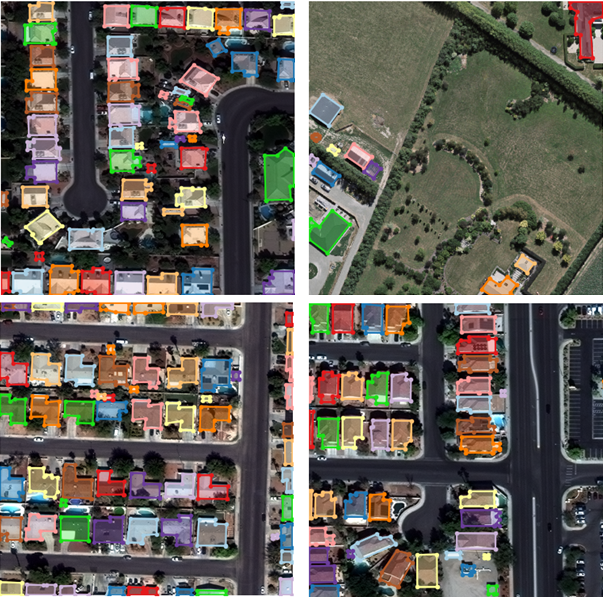}
\caption{Proximity Based Analysis}
\label{fig_3}
\end{figure}

 Figure \ref{fig_3} illustrates the model's ability to accurately segment individual buildings, even in cases where structures share corners or lie along image boundaries. Decoupled-PolyGCN excels in resolving issues of overlapping boundaries, capturing distinct building shapes with high precision. By incorporating orientation information and utilizing the CO-DETR detection network, the model ensures robust performance in complex and dense urban settings. These advancements highlight its reliability and adaptability for real-world applications, effectively addressing the challenges posed by dense urban environments and ensuring accurate segmentation of individual structures.
\newline \textbf{Incomplete or Erroneous Annotations:} A recurring challenge encountered during our experiments was the presence of very small, truncated buildings near image boundaries. These structures often caused performance degradation due to incomplete or inaccurate annotations, as both the network and human annotators struggled to define the shapes of such buildings correctly. To assess the impact of these problematic annotations, we conducted experiments by excluding truncated buildings from the evaluation process. This exclusion led to noticeable improvements in performance metrics, including PoLiS, which measures geometric distances between predicted and ground truth polygons, as well as AP and AR scores. Additionally, errors often occurred when a large building in close proximity to a very small building was incorrectly annotated as a single structure in the ground truth. Despite these challenges, the results presented in all tables within this thesis include evaluations conducted with the incorrect annotations to ensure fair comparisons across all models and scenarios. Notably, our final model, Decoupled-PolyGCN, exhibited remarkable robustness, accurately predicting many building footprints even when the ground truth annotations were imperfect. This underscores the model's reliability and adaptability in handling challenging and inaccurately annotated datasets.

\par We performed extensive analyses and experiments across multiple datasets, showcasing the progressive evolution of our models, which culminated in the highly robust and accurate Decoupled-PolyGCN. Starting with the baseline model, R-PolyGCN, we systematically identified its limitations and addressed them through iterative improvements. Each enhancement incorporated innovative modules and advanced techniques, progressively increasing the network's ability to extract building footprints with high accuracy while maintaining geometric regularity.

\par Key advancements included the integration of orthogonality loss, the Feature Augmentation Module (FAM), Attraction Field Maps (AFMs), and a decoupled object detection strategy. These improvements enabled the network to overcome challenges associated with varying building sizes, complex geometries, and densely packed urban environments. The final model, Decoupled-PolyGCN, outperformed existing state-of-the-art methods, demonstrating its superior capability in handling diverse urban and rural scenarios. This systematic progression and iterative refinement have resulted in a robust and efficient framework for building footprint extraction. The advancements introduced by Decoupled-PolyGCN pave the way for future innovations in urban mapping applications and remote sensing technologies, highlighting its potential for real-world implementation and impact.
\section{Conclusion}
Our model, Decoupled-PolyGCN, employed a two-step approach to address the limitations of earlier designs. First, it utilized CO-DETR, a state-of-the-art transformer-based detection network, to generate high-quality bounding boxes by decoupling object detection from building polygon segmentation. These bounding boxes were then processed by a UResNet101 backbone, which provided multi-scale features essential for accurate polygon reconstruction. The Graph Convolutional Network (GCN) iteratively refined these polygons, incorporating dynamic point initialization and orthogonality loss to produce regularized and geometrically precise building boundaries. This design enhanced segmentation accuracy and significantly reduced inference time, showcasing scalability and efficiency for real-world applications. Both quantitative and qualitative evaluations validated the effectiveness of our proposed methods. Decoupled-PolyGCN consistently outperformed the state-of-the-art HiSup model, achieving a 6\% improvement in Average Precision (AP) and a 10\% improvement in Average Recall (AR) on the WHU dataset. On the SpaceNet-2 dataset, the model delivered a 5\% improvement in AP and a 7\% improvement in AR. Qualitative results demonstrated the model's robustness across diverse scenarios, including varying building sizes, complex geometries, closely packed structures, and occluded buildings. The advancements introduced by Decoupled-PolyGCN highlight its potential as a scalable, efficient, and reliable solution for automated building footprint extraction in real-world applications.

\section*{Acknowledgments}
This should be a simple paragraph



%

\section{Biography Section}

\vspace{11pt}

\vspace{-33pt}
\begin{IEEEbiography}[{\includegraphics[width=1in,height=1.25in,clip,keepaspectratio]{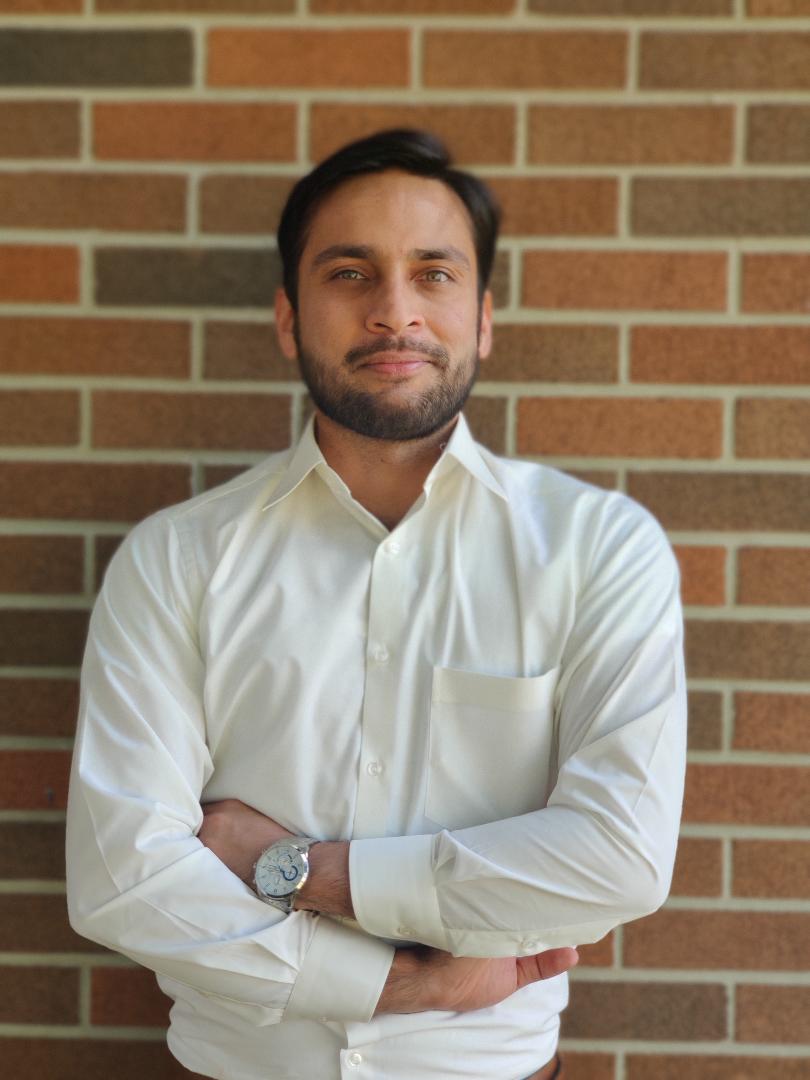}}]{Muhammad Kamran} holds a PhD from York University in Earth and Space Science and Engineering. He has a background in AI and Machine Learning and is focused on research in that domain.
\end{IEEEbiography}

\vfill

\end{document}